\documentclass[%
 aip,
 amsmath,amssymb,
 reprint,%
]{revtex4-1}

\usepackage{graphicx}
\usepackage{dcolumn}
\usepackage{bm}

\usepackage[utf8]{inputenc}
\usepackage[T1]{fontenc}
\usepackage{mathptmx}
\usepackage{etoolbox}
\usepackage{color}
\usepackage{footnote}

\makeatletter
\def\@email#1#2{%
 \endgroup
 \patchcmd{\titleblock@produce}
  {\frontmatter@RRAPformat}
  {\frontmatter@RRAPformat{\produce@RRAP{*#1\href{mailto:#2}{#2}}}\frontmatter@RRAPformat}
  {}{}
}%
\makeatother
\begin{document}

\preprint{AIP/123-QED}

\title{Enhancing Efficiency and Propulsion in Bio-mimetic Robotic Fish through End-to-End Deep Reinforcement Learning}

\author{Xinyu Cui\textsuperscript{\dag}}
\affiliation{Institute of Automation, Chinese Academy of Science, Beijing 100190, China}
\affiliation{School of Artificial Intelligence, University of Chinese Academy of Sciences, Beijing 100049, China}

\author{Boai Sun\textsuperscript{\dag}}
\affiliation{ 
Zhejiang University-Westlake University Joint Training, Zhejiang University, Hangzhou 310027, China
}%
\affiliation{Key Laboratory of Coastal Environment and Resources of Zhejiang Province, School of Engineering, 
Westlake University, Hangzhou 310030, China}

\author{Yi Zhu}
\affiliation{Key Laboratory of Coastal Environment and Resources of Zhejiang Province, School of Engineering, Westlake University,
Hangzhou 310030, China}
\affiliation{Institute of Advanced Technology, Westlake Institute for Advanced Study, Hangzhou 310024, China}

\author{Ning Yang*}
\affiliation{Institute of Automation, Chinese Academy of Science, Beijing 100190, China}
\affiliation{School of Artificial Intelligence, University of Chinese Academy of Sciences, Beijing 100049, China}

\author{Haifeng Zhang*}
\affiliation{Institute of Automation, Chinese Academy of Science, Beijing 100190, China}
\affiliation{School of Artificial Intelligence, University of Chinese Academy of Sciences, Beijing 100049, China}

\author{Weicheng Cui* }
\affiliation{Key Laboratory of Coastal Environment and Resources of Zhejiang Province, School of Engineering, Westlake University,
Hangzhou 310030, China}
\affiliation{Institute of Advanced Technology, Westlake Institute for Advanced Study, Hangzhou 310024, China}

\author{Dixia Fan* }
\affiliation{Key Laboratory of Coastal Environment and Resources of Zhejiang Province, School of Engineering, Westlake University,
Hangzhou 310030, China}
\affiliation{Institute of Advanced Technology, Westlake Institute for Advanced Study, Hangzhou 310024, China}

\author{Jun Wang}
\affiliation{Department of Computer Science, University College London, London WC1E 6BT, United Kingdom}

\collaboration{\textsuperscript{\dag} Equal contribution; *Authors to whom correspondence should be addressed: ning.yang@ia.ac.cn; haifeng.zhang@ia.ac.cn; cuiweicheng@westlake.edu.cn; fandixia@westlake.edu.cn; }
\footnote{Authors to whom correspondence should be addressed: email 1: ning.yang@ia.ac.cn; email 2: haifeng.zhang@ia.ac.cn; email 3: cuiweicheng@westlake.edu.cn; email 4: fandixia@westlake.edu.cn; }


\begin{abstract}
Aquatic organisms are known for their ability to generate efficient propulsion with low energy expenditure. While existing research has sought to leverage bio-inspired structures to reduce energy costs in underwater robotics, the crucial role of control policies in enhancing efficiency has often been overlooked. In this study, we optimize the motion of a bio-mimetic robotic fish using deep reinforcement learning (DRL) to maximize propulsion efficiency and minimize energy consumption. Our novel DRL approach incorporates extended pressure perception, a transformer model processing sequences of observations, and a policy transfer scheme. Notably, significantly improved training stability and speed within our approach allow for end-to-end training of the robotic fish. This enables agiler responses to hydrodynamic environments and possesses greater optimization potential compared to pre-defined motion pattern controls. Our experiments are conducted on a serially connected rigid robotic fish in a free stream with a Reynolds number of 6000 using computational fluid dynamics (CFD) simulations. The DRL-trained policies yield impressive results, demonstrating both high efficiency and propulsion. The policies also showcase the agent's embodiment, skillfully utilizing its body structure and engaging with surrounding fluid dynamics, as revealed through flow analysis. This study provides valuable insights into the bio-mimetic underwater robots optimization through DRL training, capitalizing on their structural advantages, and ultimately contributing to more efficient underwater propulsion systems.
\end{abstract}
\maketitle

\section{Introduction}

The control of propulsion in fluids has undergone extensive development in human history, with a multitude of methods being employed, such as paddling, jetting, and propeller-based propulsion systems\cite{sun2022recent}. 
Many motion patterns mimic the reciprocating structure, leveraging wheel-like mechanical properties. However, they often lack efficiency, maneuverability, and noise control compared to natural fish-like aquatic organisms.
In recent years, researchers have been delving into the simulation of advanced structures and motion control modes of fish, demonstrating exceptional hydrodynamic performance due to their multi-link coordinated undulation\cite{WOS:001035069000001}. Unlike the significant turbulence generated in the wake by propeller propulsion modes, the bodily waves of fish interact with the flow field at varying positions along their anterior and posterior edges, maximizing the utilization of the flow field’s energy.

Given that the propulsion structures of fish comprise flexible bodies, theoretical simulations and the design of robotic fish often adopt a segmented multi-link structure for actuation in an effort to simplify the model\cite{WOS:000276556300006, wang2022research,korkmaz2015dynamic}. This facilitates the optimization of thrust and energy efficiency for bio-mimetic robotic fish, aiming to emulate the superior performance of natural fish\cite{WOS:000600804300001}. However, even with the simplified model of rigid linkages, controlling the independent links introduces a considerable number of parameters\cite{WOS:001100758500001}. 
Furthermore, fish exhibit diverse range of motion patterns, including conventional head-first swimming and tail-first swimming modes\cite{dhileep2023investigation}. This variation in swimming behavior adds another layer of challenge to the  control optimization process. Therefore, optimization of the control structure often involves combining trigonometric functions of the same frequency for different angular parameters within each cycle\cite{WOS:000942071400015, WOS:000244117400072, WOS:000351205700010, WuZ2012, WOS:000326358700003, YuJ2015, WangM2019}. 

With the rapid advancement of artificial intelligence, researchers are turning to deep reinforcement learning (DRL) techniques\cite{Arulkumaran2017,colabrese2017flow} to refine control modes in enhancing robotic fish\cite{YanS2020, Hu2023} performance. These studies often employ DRL algorithms to optimize the trigonometric functions' parameters.
Nevertheless, fish in nature exhibit complex motion patterns that simplified models like trigonometric functions can capture. Adhering strictly to a predefined trigonometric swimming pattern hinders immediate and optimal response to fluid dynamics, posing challenges in leveraging hydrodynamics and embodiment. 
To maximize bio-mimetic robotic fish's hydrodynamic potential, it is crucial to liberate robotic fish from fixed parametric patterns, enabling them to make distinct motion decisions at each step.
Instead of relying solely on trigonometric motions, we leverage DRL's decision-making capabilities to optimize a bio-mimetic robotic fish's motion patterns without predefined functions' constraints. This involves training the robotic fish in a computational fluid dynamics (CFD)-based simulation environment in an end-to-end manner. With the adoption of an end-to-end scheme, significant instability emerges. To counter this, we implement multiple approaches to bolster DRL's performance and mitigate instability.

In this work, we introduce a novel DRL approach that incorporates the following four innovative aspects.
First, inspired by the way fish perceive local velocity and pressure through their lateral line, we integrate flow pressure sensors around the robotic fish's surface to enhance the perception and utilization of water flow. Second, to address the constraints of single-step decision-making within limited environmental observations, we integrate sequential past observations and a self-attention mechanism. This enhancement aims to grant the DRL agent a more comprehensive understanding of its environment. Next, we pre-train the DRL agent with knowledge of trigonometric swimming motions, which significantly expedites the training process. Lastly, we employ clipping functions to process raw data and filter out invalid information from the simulator while gently discouraging the agent from taking actions with high energy costs.


The rest of this paper is organized as follows. The numerical simulation models and DRL method are presented and explained in Section \ref{sec:method}. The DRL result overview, the policy optimization process, the comparison between featured result cases, and the analysis of the high-efficiency motion pattern are illustrated in Section \ref{sec:result}. Finally, concluding remarks are addressed in Section \ref{sec:conclusion}.

\section{Methodology}\label{sec:method}
\subsection{Numerical Simulation Model}
\begin{figure*}[ht]
    \centering
    \includegraphics[width=1\linewidth]{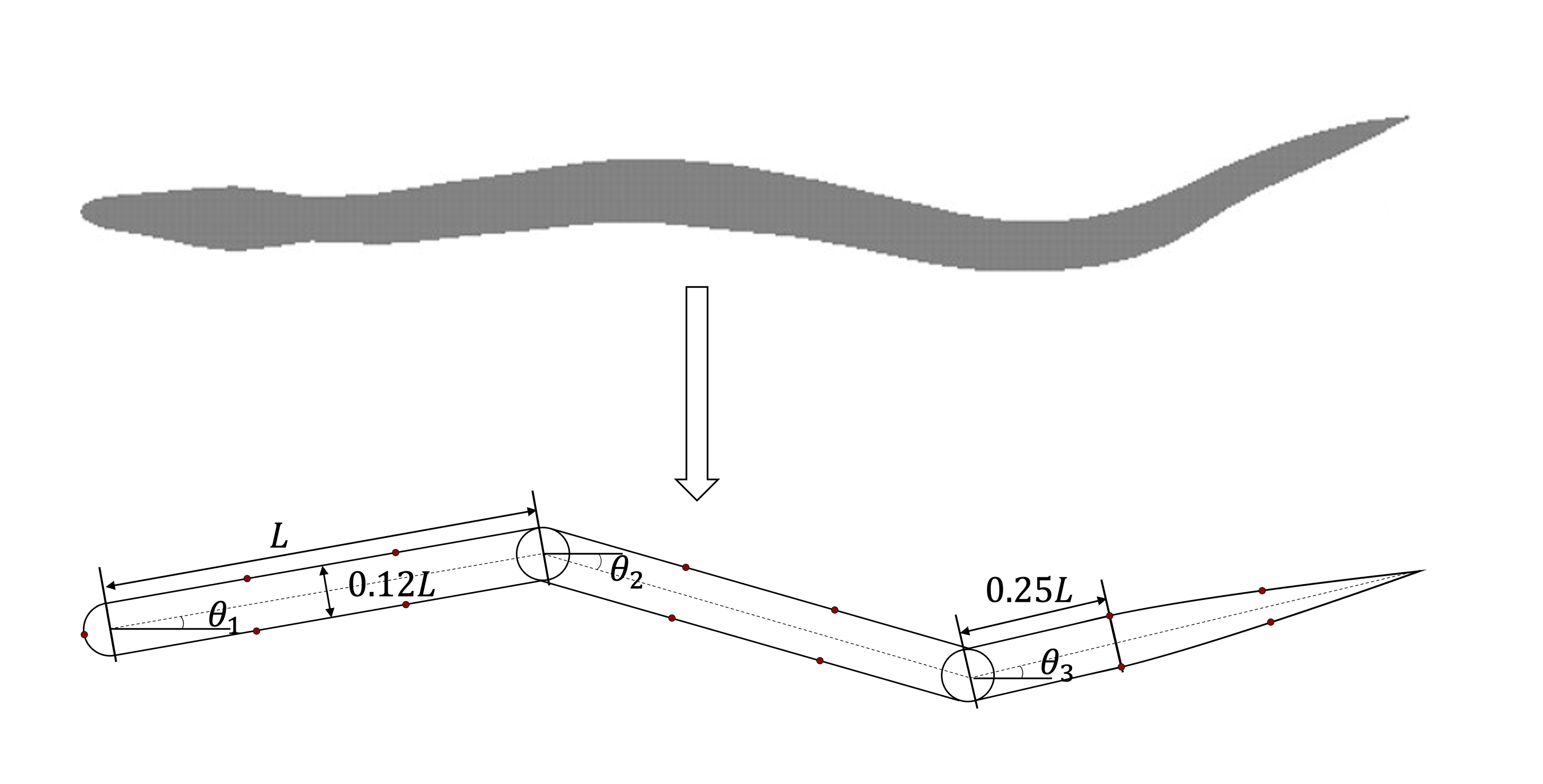}
    \caption{The numerical structure of the three-joint fish. Here we implement a part of the NACA0012 foil at the last joint to better imitate a fish. The red dots represent flow pressure sensors set around the robotic fish's surface.}
    \label{fig:fish3}
\end{figure*}

The propulsion of aquatic organisms can be modeled as a composition of rotation and translation of rigid body links serially connected by different links. For simplicity and generalizability, a three-link fish with 4 degrees of freedom (DOFs) is numerically investigated in a two-dimensional flow field.  As shown in Fig.~\ref{fig:fish3}, the three links resembling the head, trunk, and tail are of the same length and serially connected by joint 0 and joint 1, and move in a given pattern in the flow field. To better simulate the fluid mechanics of fish, the posterior part of a standard NACA airfoil is implemented at the last link of the model.

The numerical experiment is conducted in a uniform two-dimensional flow field, where a steady flow to the right is introduced at the left edge. The three angles define the motions of the links to the horizontal of the links respectively, and the overall motion of the first connection point in the y-direction, perpendicular to the direction of the fluid flow. By assuming the fish obtains a steady swimming pattern that approximately achieves quasi-static progress in the incoming stream, we fix the x-direction motion of the fish. The remaining four parameters {$\theta_{1}(t), \theta_{2}(t), \theta_{3}(t), y(t)$} that vary with time form a complete description of the overall complex motion. $\theta_{i}(t),i\in[1,2,3]$ defines the angles of each link relative to the horizontal direction, and $y(t)$ stands for the overall y-directional position of the fish, with the reference point fixed at the end of the first link.

In our study, the Reynolds number is defined as :
\begin{equation*}
    Re=\frac{\rho U_{\infty} L^{*}}{\mu},
\end{equation*}
and is set to 6000, where $\rho$ is the fluid density, $U_{\infty}$ is the inflow velocity from the left side, and $\mu$ is the dynamic viscosity.
Considering the complex structure of the fish, we select $L^*=3L$ as the characteristic length, where $L$ is the length of each part. The Strouhal number $St$ of the motion is defined as : 
\begin{equation*}
    St=\frac{ L^{*}}{T U_{\infty}}.
\end{equation*}

For trigonometrical motions, we set the frequency of the three links as the same for a steady swimming pattern, and the period is defined as $T$. After the optimization of DRL, the output motion policies are no longer parametric, yet all the results still present periodicity (as shown in Fig.~\ref{fig:iteration}). Thus, we intercept time periods in different results to evaluate the $St$ and other quantities that require averaging over time.

Several important non-dimensional coefficients are used to describe the overall hydrodynamic performance of the three-link fish, including the thrust coefficient $C_T$, the lateral force coefficient $C_L$, the power consumption coefficient $C_P$, and the propulsion efficiency $\eta$. The thrust coefficient $C_T$ is defined as follows, 
\begin{equation}
\left\{
\begin{aligned}
&C_T=\frac{\overline{F_x}}{1/2 \rho U_{\infty}^2 c L}\\
&\overline{F_x}=\frac{1}{\tau} \int_0^\tau F_x(t) \delta t ,
\end{aligned}\right. 
\end{equation}
where $\overline{F_x}$ stands for the mean force in the x-direction, $c$ for the fish width, $L$ for the fish characteristic length, and $\tau$ for the accumulated time with stable flow field in the simulation. Similar as $C_T$, we define $C_L$ as
\begin{equation}
\left\{
\begin{aligned}
&C_L=\frac{\overline{F_y}}{1/2 \rho U_{\infty}^2 c L}\\
&\overline{F_y}=\frac{1}{\tau} \int_0^\tau F_y(t) \delta t ,
\end{aligned}\right. 
\end{equation}
where $\overline{F_y}$ represents the mean force in the y-direction. Lastly, we have $C_P$ as 
\begin{equation}
\left\{
\begin{aligned}
&C_P=\frac{\bar{P}}{1/2 \rho U_{\infty}^3 c L}\\
&\bar{P}=\frac{1}{\tau}\int_0^\tau \sum_{i=1,2,3} \left( M_{i}(t) \dot\theta_{i}(t) + F_{i}(t) \dot{\vec{r}}_{i} (t) \right) \delta t ,
\end{aligned}\right. 
\end{equation}
where $\bar{P}$ stands for time-averaged power consumption, and $M_{\theta i}(t)$ for the pitching torque of each link from the fluid. The overall performance of the three-link fish is to be evaluated by multiple characteristics such as propulsion ability, swimming stability and propulsion efficiency. Therefore, the reward of the optimization algorithm is defined as a complex compound function considering all the coefficients. 

The boundary data immersion method (BDIM) represents a novel approach to handling boundary conditions in fluid-solid interaction problems, merging robustness and precision as evidenced in previous studies \cite{maertens2015accurate, weymouth2011boundary, weymouth2006advancements}. This method simplifies the complex multi-domain problems by integrating them into a single domain with a smoothly immersed interface.

Essentially, BDIM is an innovative take on the immersed boundary method (IBM). The classic conformal grid method, such as the Finite Volume Method\cite{moukalled2016finite} used in FLUENT, differentiates solid from fluid domains, simplifying the computation of fluid domain mesh. However, this mesh is irregular and requires frequent updates, leading to high computational demands. This complexity is particularly notable when dealing with sharp edges or deformable objects. On the other hand, the classic immersed boundary method \cite{huang2019recent} facilitates a uniform mesh but adjusts fluid motion at multi-phase boundaries, like rigid body surfaces, through distributed "bulk forces". This approach, while simplifying the mesh generation, compromises on higher-order hydrodynamic terms, thereby reducing accuracy. 

The BDIM, in contrast to the two methods previously discussed, effectively merges the distinct domains of rigid bodies ($B$), fluid ($F$), and the boundary layer ($S$) into a unified field, denoted as $\Omega$. This approach employs a finite-width smoothing kernel at the boundary, applying the solid object's boundary conditions onto the fluid domain. This ensures adherence to the no-slip condition at the surface of the object, integrating the domains seamlessly, and achieving a good balance between computational speed and accuracy. In this study, we perform the CFD solver implementing the BDIM algorithm, along with the computation of the inverse dynamics of the simulated fish model on the processing platform. The validation of the algorithm is shown in Appendix \ref{app:BDIM}.

\subsection{Trigonometrical Motions and Data Collection}

The traditional control method for serially connected robotic fish involves using multiple trigonometric curves to represent rotation patterns across multiple links. This method can be defined as a trigonometrical swimming policy for robotic fish. To apply the control policy, we define four actions$[a_1, a_2, a_3, a_4]$ for the simulated robotic fish's 4 DOFs: the rotation of head to joint 0, the rotation of trunk to joint 0, the rotation of tail to joint 1, and movement in y-dimension of joint 0. The action space ranges from $-0.03$ to $0.03$ for each DOF, allowing a maximum change of 0.03 at each simulation step. Utilizing this trigonometric policy, a swimming pattern characterized by a cosine curve can be defined as:

\begin{equation}
\left\{
\begin{aligned}
    t_n &= \sum_{j=0}^n \delta t_j \\
    a_i &= A_i cos( \omega_i t_n + p_i) , i \in [1,2,3,4] .
\end{aligned}\right.  
\end{equation}
Here, $t_n$ represents the simulation time till the simulation step $n$. The actions, as mentioned earlier, $a_1$, $a_2$, and $a_3$ correspond to the rotational movements of the links at simulation time step $n$, while $a_4$ controls the vertical movement of the entire robotic fish along the y-dimension. For the parameters, $A_i$ determines the amplitudes of the actions, $\omega_i$ determines the circular frequency, and $p_i$ determines the relative phases. 

To optimize the parametric trigonometrical policy of the simulated robotic fish, a brute force\cite{Heule2017} algorithm is adopted. This algorithm facilitates the search for the optimal trigonometric policy by tuning parameters $A_i$, $\omega_i$, and $p_i$.
Through the brute force search, the Pareto frontier between the efficiency and propulsive capacity of the swimming gaits produced by trigonometrical  policy is approximately ascertained. The threshold of propulsive capacity delineated by the trigonometrical  policy can be established as a baseline. During this phase, an assortment of offline data has been gathered, encompassing link angular configurations, angular velocities, and pressure distributions in the proximity of the simulated robot. This repository of offline data enables the deployment of a transfer learning methodology for pre-training a basic policy in subsequent DRL experiments.


\subsection{Markov Decision Process Formulation and DRL Algorithm}
In aforementioned experiments, we initially optimize the swimming policy of the simulated robotic fish in a trigonometric manner. However, real aquatic organisms exhibit more complex motions beyond what trigonometric curves can capture. To break away from dependence on predefined trigonometric curves, our goal is to optimize the swimming policy in an end-to-end fashion. We formulate the swimming of the robot as a Markov decision process (MDP) and optimize it through a DRL approach. The MDP is characterized by a tuple $\langle S, A, P, R, \gamma \rangle$, where $S$ represents the set of states observed by the agent, $A$ represents the set of actions chosen by the agent, $P$ represents the deterministic transition function that defines the state transition dynamics from step $n$ to $n+1$ as $s_{n+1} = P(s_n, a_n)$. The function $R(s, a)$ represents the reward function, which returns a scalar reward at step $n$, denoted by $r_n \sim R(s_n, a_n)$. Lastly, $\gamma\in [0,1]$ is the discount factor. The goal of DRL is to find the optimal policy $\pi^*: S \rightarrow A$ that maximizes the expected return (cumulative discounted reward):
\begin{equation}
    \pi^* := \underset{\pi}{\text{arg max}} \mathbb{E}_{s_{n+1} = P(s_n,a_n), a_n \sim \pi(s_n)} \left[\sum_{n=0}^{\infty} \gamma^n r_n \right] .
\end{equation}

To provide the agent with environmental information resembling that experienced by real fish, the agent's state $s_n$ at step $n$ is compiled as a one-dimensional array: $[\theta_i, \omega_i, y, SP]$. This array concatenates angles of three links around joints, angular velocities of links, the y-coordinate, and a sequence of flow pressure collected from thirteen evenly distributed surrounding points in the robot's surface.
The agent's actions align with the trigonometric policy definition.
For enhancing thrust while minimizing energy consumption, the reward function $r_n$ is defined as: 
\begin{equation}\label{eq:rewardfunc}
    r_n = \alpha \cdot 75 \cdot symlog (\frac{{I_x}_n}{75}) - \beta \cdot 1000 \cdot symlog (\frac{W_n}{1000}) + \xi_n .
\end{equation}
Here, ${I_x}_n \in [-150, 150]$ represents the impulse in the x-dimension obtained by the agent at step $n$, $W_n \in [-2000, 2000]$ signifies the energy consumed by the agent at step $n$, and $\xi_n$ denotes the penalty at step $n$. The detailed penalty function setting can be found in Appendix \ref{app:penalty}. Hyper-parameters $\alpha$ and $\beta$ govern the trade-off between impulse and energy consumption within the reward structure. The $symlog$ function exhibits linear behavior between 0 and 1, and logarithmic behavior after 1, thus playing a role on softly clipping unstable values over the set limits: 
\begin{equation}
symlog(x) =  sign(x) \cdot log (|x| + 1) .
\end{equation}

\begin{figure*}[ht]
    \centering
    \includegraphics[width=1\textwidth]{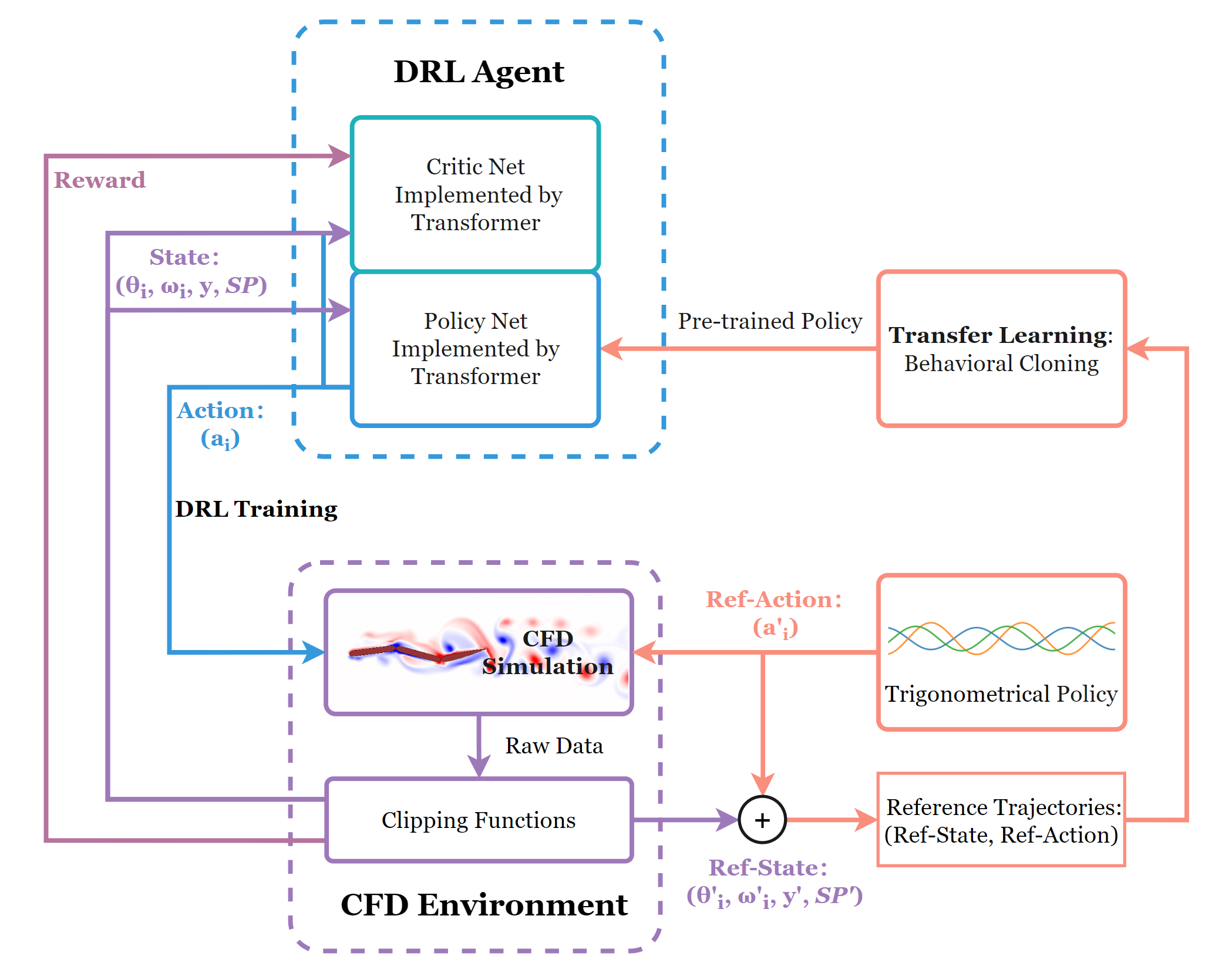}
    \caption{Framework of our training pipeline. At the beginning, we pre-train the action policy with quantities of reference trajectories collected by selected trigonometrical policies. Then during DRL training stage, the agent receive processed data from CFD environment and make a decision at each simulation step. The policy is trained and updated with data collected from each episode at the end of it.}
    \label{fig:Algo_structure}
\end{figure*}

To optimize this problem, we employ the proximal policy optimization (PPO)\cite{Schulman2017} algorithm as the DRL framework and our training pipeline is illustrated in Fig.~\ref{fig:Algo_structure}. 
PPO adopts the actor-critic framework in DRL, which comprises two essential components: policy network $\pi(a|s)$ parameterized by $\theta$ and value network $V(s)$ parameterized by $\phi$. The policy network generates actions based on the states for the DRL agent to engage with the environment. The loss function of policy network is defined as:

\begin{equation}
    L_{\text{actor}}(\theta) = \hat{\mathbb{E}}_n\left[\min\left(\rho_n(\theta)\hat{A}_n, \operatorname{clip}\left(\rho_n(\theta), 1-\epsilon, 1+\epsilon\right)\hat{A}_n\right)\right] . \label{actorloss} 
\end{equation}

Here, $\rho_n(\theta)$ represents the probability ratio of policies, PPO clips this ratio by hyper-parameter $\epsilon$ to ensure a stable improvement. $\hat{A}_n$ is  advantage function that reduces policy gradient variance. 

The value network $V(s)$ is trained to predict the value of current state, representing an estimate of the cumulative rewards experienced thus far. This network aids the DRL agent's learning process by contributing to the estimation of advantage. The loss function of the critic network is designed to minimize temporal difference loss:

\begin{equation}
    L_{\text{critic}}(\phi) = r_n + \gamma V\left(s_{n+1};\phi\right) - V\left(s_n;\phi\right) .
\end{equation}

The comprehensive loss function for the PPO algorithm can be expressed as follows, and it can be optimized using gradient-based methods:

\begin{equation}
    L_{\text{PPO}}(\theta, \phi) = L_{\text{actor}}(\theta) + L_{\text{critic}}(\phi) .
\end{equation}

PPO leverages the trust region optimization approach, employing an efficient estimation detailed in equation \eqref{actorloss}.
This method restrains policy changes during updates, ensuring proximity between the new and previous policies while enabling substantial improvements. It guarantees consistent policy enhancement even in scenarios with incomplete or noisy states, which is suitable for tasks in  CFD environments.
The performance of PPO algorithm is further improved by adopting two other approaches, we adopt generalized advantage estimation (GAE)\cite{Schulman2015} and normalized rewards for optimizing value and advantage estimations.
GAE balances the trade-off between bias and variance, facilitates faster learning and requires fewer samples for policy optimization. Normalized rewards lead to more stable gradients and also facilitate faster learning and more stable performances.

\subsection{Task-Specific Approaches for Enhancing DRL Performance}

Numerous studies have focused on optimizing robotic fish swimming policies. However, many have chosen indirect control methods such as employing trigonometric curves to address training instability. But this results in an incomplete search space for the gaits of the robotic fish. To the best of our knowledge, we are the first ones to optimize robotic fish's swimming strategy within a CFD environment using an end-to-end scheme. 
However, challenges emerge due to this approach, directly incorporating DRL fails to formulate an effective swimming strategy, letting alone optimize it. To overcome these hurdles, we adopt several task-specific approaches to enhance DRL performance.

\textbf{Extended Observations.} Firstly, the agent may struggle to achieve an optimal policy due to the limited observable information. Wang et al. utilized on-board pressure sensors to enhance a robotic fish's perception and aid its navigation along a wall\cite{WangW2017}. Nevertheless, studies optimizing robotic fish motion only offer observations on kinematic and positional information to the agent. To overcome this limitation, we leverage our CFD environment to calculate the flow pressure along the robotic fish's surface. By incorporating processed pressure data into the agent's observations, the agent's perception of the hydrodynamic environment is extended.

\textbf{Transformers with sequential information.} Secondly, it is necessary to tackle the instability arising from single-step decision making in an environment with limited observations. 
In nature, fish utilize memory to better understand their environment and make decisions. Joonho Lee, et al demonstrated that incorporating a short history of observations can enhance environment perception\cite{JoonhoL2020}. Inspired by this, we adopt a transformer\cite{Vaswani2017}  with a sequence of past
observations as input to replace the multi-layer perception (MLP) network. 

\textbf{Transferring trigonometrical swimming knowledge.} Thirdly, exploring optimal swimming strategies involves adopting an end-to-end control method, which offers a vast state-action space but poses optimization challenges. It is time-consuming for DRL to optimize  from scratch.
To tackle with this problem, the technique of transfer learning\cite{Weiss2016}\cite{HuangB2021} helps. Specifically,  we endeavor to transfer the knowledge of trigonometric swimming policies to the DRL agent's initial policy. In practice, we implement a behavior cloning (BC) scheme\cite{Bain1995} to align the initial policy's behavior with that of trigonometric policies. BC is a simple imitation learning\cite{Hussein2017} method, which directly using reference trajectory data which contains reference state-action pairs $({s_{ref}}, {a_{ref}})$, and minimizing the difference between the reference action ${a_{ref}}$ and the policy network's output $\pi_{\theta}({s_{ref}})$. For example, a mean squared error (MSE) loss function of BC can be defined as:
\begin{equation}
    L_{\text{BC}}(\theta) = \frac{1}{N} \sum_{i=0}^N({a_{ref}}_i - \pi_{\theta}({s_{ref}}_i))^2 .
\end{equation}
Subsequently, a gradient-based pre-training method can be employed to minimize the discrepancy between reference actions and the actions generated by the policy network. Following the pre-training, the policy can be integrated into the PPO framework for further optimization.

\textbf{Symlog Clipping Functions and Early Stopping.} To encourage the agent to explore, the agent is able to achieve any position within its 4 DOFs. However, the agent may attain invalid positions that are impossible in reality, causing instability in DRL training. 
Peng, et al proved that training with invalid data will decrease the policy's performance \cite{PengX2018}. To avoid these situations, we terminate simulations and give the agent a punishment in reward whenever the agent reaches invalid positions, gets any invalid angular velocities, or makes the fluid field chaotic. Additionally, to improve the stability throughout training, $symlog$ functions are incorporated into reward function. These functions filter invalid reward signals and softly regulate the agent's pursuit of excessively high impulses, which consume excessive energy.

\section{Analysis and Discussion}\label{sec:result}
\subsection{Ablation Study on Task-Specific Approaches}

\begin{figure*}[htbp!]  
    \includegraphics[width=\textwidth]{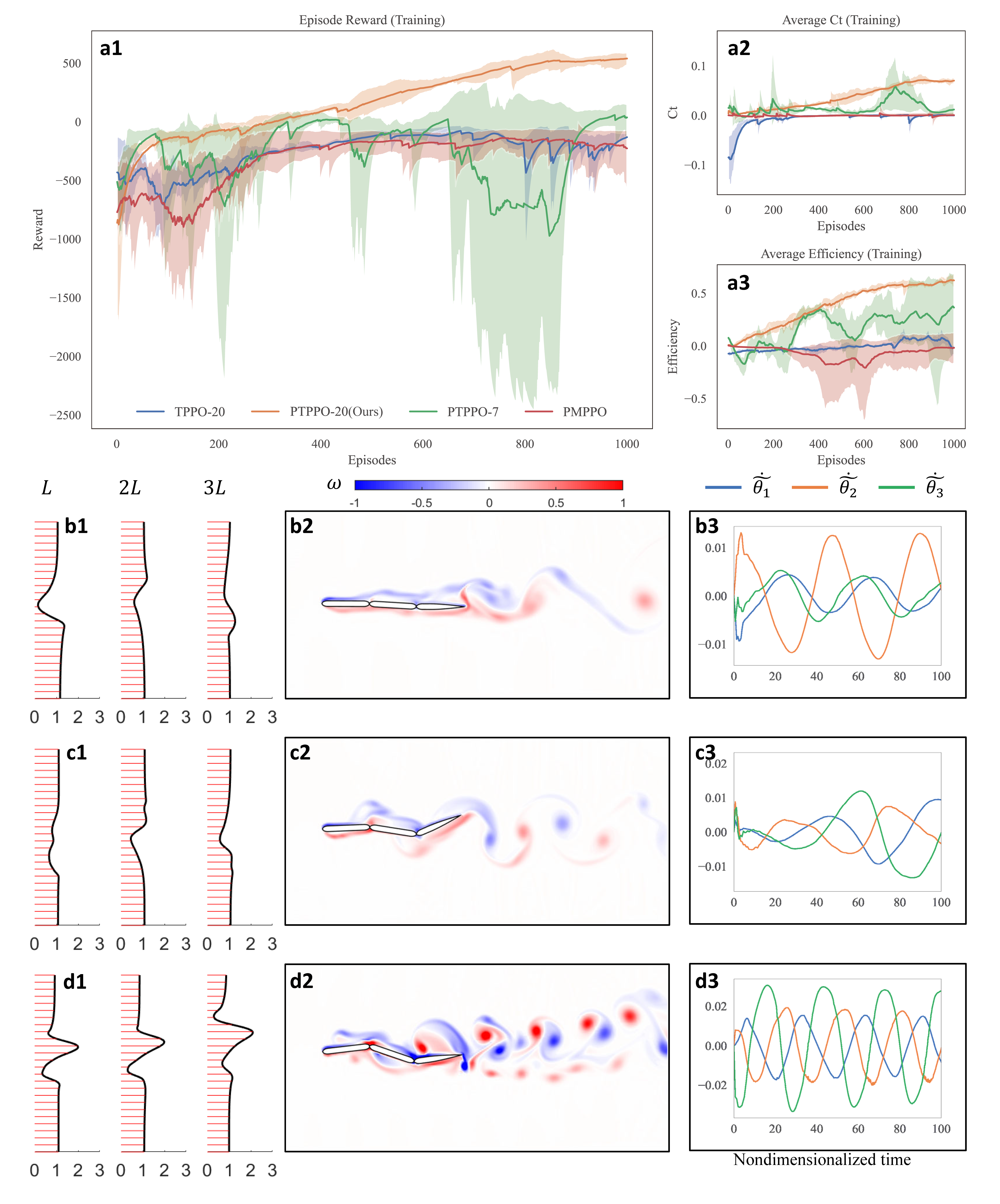}
    \caption{(a) Comparison among four scenarios. (a1) The episode reward plot throughout the training. (a2) The average thrust indicator $C_T$ of each episode throughout the training. (a3) The average efficiency of each episode throughout the training. For each baseline, we train $5$ different models varying only over random seeds. We plot the average performance as the solid line, and the shaded area represents the performance boundaries between the upper and lower limits. Performance metrics like $C_T$ and efficiency were measured during the training process, due to the action noise introduced during DRL training, the observed performance might be slightly lower than that achieved during actual execution.
    (b-d)The result of DRL's policy evolution: An example of a high-efficiency case in the evolution of DRL policies throughout the training, representing the starting, middle (200 episodes), and ending (1000 episodes) stages. (b1-d1) The x-directional flow velocity distribution along the vertical lines at distances of 1L, 2L, and 3L from the tail. (b2-d2) The vorticity field generated by the fish's motion. (b3-d3) The evolution of angular velocities of the three links.}
    \label{fig:iteration}
\end{figure*}

To validate the task-specific approaches' effectiveness, a comparative training and analysis among four scenarios is conducted: \textbf{PTPPO-20 (Our method)}: PPO utilizing pre-trained networks implemented by transformers while considering pressures; \textbf{PTPPO-7}: PPO utilizing pre-trained networks implemented by transformers without considering pressures; \textbf{PMPPO}: PPO utilizing pre-trained networks implemented by MLPs while considering pressures, and \textbf{TPPO-20}: PPO with non-pre-trained networks implemented by transformers while considering pressures. We take the episode reward, $C_T$, and relative efficiency as three evaluation indicators, and the results are plotted in Fig.~\ref{fig:iteration}.

\textbf{Adequacy of Kinematic Information:} Pre-training coupled with a sequence model implemented by transformers allows the agent to steadily enhance its thrust and efficiency. However, the absence of pressure information may compromise the accuracy of policy gradients or value functions. As shown in sub-figure \ref{fig:iteration}(a), PTPPO-7 exhibits an upper bound akin to PTPPO-20, but its training curve is considerably more unstable, potentially hindering convergence to an optimal policy.

\textbf{Effectiveness of Sequential Information:} The limited variation in single-step information across different simulation steps makes it inadequate for decision-making. As a consequence, MLPs fail to take a short-horizon observation history into consideration, it fails to find policies which can generate sufficient thrust or efficiency. 
As depicted in sub-figure \ref{fig:iteration}(a), both PTPPO-7 and PTPPO-20 outperform PMPPO by a significant margin.

\textbf{The Impact of Transfer Learning on Policy Optimization:} As shown in Fig.~\ref{fig:iteration}, it can be noticed that without pre-training, the DRL agent struggles to optimize among a vast exploration space. The non-pretrained TPPO-20 spends about 200 episodes to form forward propulsion. In contrast, the three pre-trained policies skip the period struggling to form forward propulsion, as shown in sub-figure \ref{fig:iteration}(a2).

In summary, both thrust and efficiency are improved during policy optimization, with PTPPO-20 demonstrating the most notable and stable enhancement. This result proves the efficacy of our techniques in improving swimming policies' performance, indicating the agent's improved perception of the hydrodynamic environment and utilization of acquired knowledge.

Addressing training challenges, the subsequent challenge is training different agents with varying preferences for propulsion capability or efficiency. Intuitively, the weight factors $\alpha$, $\beta$ in reward function \eqref{eq:rewardfunc} are controlled to implicitly control the preference on propulsion or efficiency.

\subsection{A Closer Look At Policy Evolution Through DRL Training}

We conduct a series of experiments, varying $\alpha$ and $\beta$ parameters to explore optimized swimming policies using the PTPPO-20 method. We select parameters favoring higher efficiency ($\alpha$ being larger), enhanced propulsion ($\beta$ being larger), or a balanced preference ($\alpha$ and $\beta$ relatively equal). The optimization process, depicted in sub-figure \ref{fig:pareto}(c), results in three distinct sets of optimized swimming policies. Notably, DRL policies operate within the same action space as trigonometrical policies. At the end of training, these DRL policies demonstrate excelling performance in both efficiency and propulsion.

In the PTPPO-20 method, the initial DRL policy is pre-trained by various trigonometric data. The pre-trained policy exhibits periodicity characteristics of trigonometric functions. However, this policy limits the full utilization of the flow-body interaction. During training, the DRL agent continuously optimizes its strategy based on the old policy, resulting in substantial improvements in episodic reward, efficiency performance, and thrust. We generate charts depicting the evolution of angular velocity curves within DRL policies in a high-efficiency case. As illustrated in the sub-figures \ref{fig:iteration}(b-d) collection, the starting (pre-trained), middle (200 episodes), and ending (1000 episodes) stages of a DRL optimization process favoring higher efficiency are selected, showcasing the fish's motion strategy at different training states. The left sub-figures \ref{fig:iteration}(b1-d1) depict the x-directional fluid velocity distribution near the trailing edge at distances of L, 2L, and 3L. The middle sub-figures \ref{fig:iteration}(b2-d2) show the stabilized flow field, while the right sub-figures \ref{fig:iteration}(b3-d3) display the angular velocities of the three links. Initially, the motion is slow with a small amplitude, and the angular velocity performs trigonometric (as shown in sub-figure \ref{fig:iteration}(b3)), leading to a gentle flow field as seen in sub-figure \ref{fig:iteration}(b2). From sub-figure \ref{fig:iteration}(b1), it is observed that the motion of the three-link fish caused a lower y-directional flow velocity near the trailing edge, indicating resistance in the flow.

As training progressed, the DRL agent attempts to modify its strategy for enhancing performance. Sub-figure \ref{fig:iteration}(c2) depicts the flow field at 200 episodes into training. At this stage, due to the ongoing updates in the DRL strategy, the fish's motion do not exhibit a strong periodicity, yet the overall resistance generated by the motion was reduced, as indicated by the flow velocity information in sub-figure \ref{fig:iteration}(c1). Upon convergence, the DRL policy gradually regains periodicity, significantly differing from the initial trigonometric approach. The motion's amplitude and frequency increased notably after DRL training. As depicted in sub-figure \ref{fig:iteration}(d1), the flow velocity near the fish's tail exceeds the incoming flow, indicating thrust generation in the fish motion. This progression highlights DRL's adaptability in fluid dynamics tasks, demonstrating how the agent harnesses information outputs from the CFD simulator. Our method showcases superior performance in swimming tasks, overcoming the constraints of trigonometric-based motions and surpassing them.

\subsection{Comparison Between two motion patterns learned by DRL}
\begin{figure*}[htbp!]
    \includegraphics[width=\textwidth]{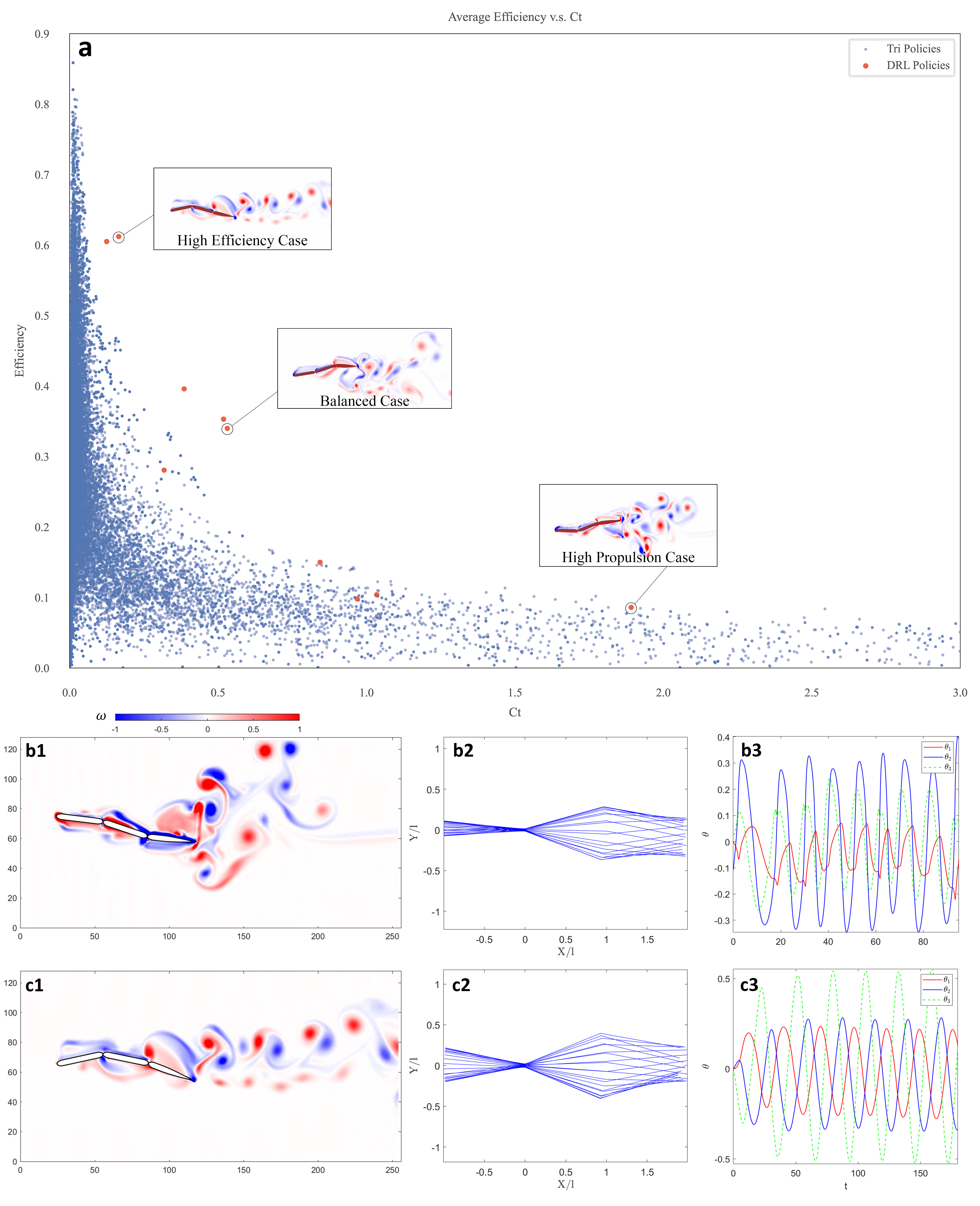}
    \caption{ (a)The result of DRL's experiments: The Pareto frontier bounded by the blue dots represents the trigonometrical policies' performance, while the red dots represent the DRL policies' performance. (b-c)The vorticity field, middle-line motion envelope, and link angles in the cases of high thrust and high efficiency, respectively. }
 \label{fig:pareto}
\end{figure*}
As shown in sub-figure \ref{fig:pareto}(a), The DRL results have generally surpassed the Brute-force results' Pareto front line in blue, and three different characteristic motion patterns emerge, emphasizing thrust, efficiency, and a balanced mode, respectively. Given that the characteristics of the balanced movement mode fall between the former two, we conduct a detailed analysis and comparison of the two distinct modal learning results. In sub-figures \ref{fig:pareto}(b)(c), the left sub-figures \ref{fig:pareto}(b1) and (c1) present the vorticity distribution of the flow field at a certain moment, while the middle sub-figures \ref{fig:pareto}(b2) and (c2) illustrate the changes in the mid-line of the fish body in the two movement modes. The right sub-figures \ref{fig:pareto}(b3) and (c3) depict the time-dependent variation functions of the angles of the three links. In each sub-figure group, the upper section corresponds to the thrust-enhanced case, and the lower section to the efficiency-enhanced case. In sub-figures \ref{fig:pareto}(b) and (c), it is evident that the vorticity in the flow field for the thrust case in sub-figure \ref{fig:pareto}(b1) is more significant (darker in color) with a more turbulent vortex structure in the tail region, whereas sub-figure \ref{fig:pareto}(c1) displays smoother vorticity with a stable shedding vortex structure at the tail. From the middle sub-figures \ref{fig:pareto}(b2) and (c2), it can be discerned that the amplitude of movement in the high thrust case is smaller with a more converged fish mid-line. This observation is further corroborated in the right sub-figures \ref{fig:pareto}(b3) and (c3). In the variation of angles over time, the efficiency-focused case exhibits a larger amplitude with its tail angle reaching up to 0.8 radians, almost double that of the thrust case. Conversely, the body swing frequency in the efficiency case is lower than in the thrust case, being nearly half of the latter. Finally, the efficiency case displays pronounced periodicity, while the movement in the thrust case appears more aggressive and disordered, with diminished periodicity. In summary, we infer that the results from DRL present two distinct modalities: one with a large swing amplitude and low frequency, offering high efficiency and almost quasi-static movement in the incoming flow (with a near-positive $C_T$ value), akin to a uniform cruising in water. In contrast, the other mode involves small amplitude, high-frequency motion, and less efficiency but generates substantial thrust, maintaining a considerable $C_T$ in the incoming flow and facilitating abrupt accelerations. This process doesn't represent a steady periodic swimming state, thus not exhibiting strong periodicity. Notably, the swing angles of both differ from the original sinusoidal swimming mode and manifest a logarithmic DRL curve of rising and falling. Under such a movement modality, when the fish body begins to sweep from one side to the other, it attains a higher speed, thus yielding increased thrust and efficiency. The uniqueness of our learned movement pattern compared to sinusoidal swimming will be further elaborated and contrasted in the subsequent section.


\subsection{Analysis of the high $\eta$ case}

From two different cases, we can compare the vorticity and pressure distribution in the flow field under high-efficiency and high-thrust conditions, and infer the relationship between good hydrodynamic performance and the interaction between fish body motion and fluid. In the case emphasizing thrust, it is observed that two sets of diverging vortices emerge on both sides of the tail due to significant and drastic swinging. The corresponding rear 1L average flow speed is also larger. In the high-efficiency case, the flow field is more gentle, and there is a smooth vortex shedding at the tail, with a smaller protrusion in the middle of the 1L velocity line. Since generating significant thrust with high frequency and large amplitude motion is relatively trivial, we focus on analyzing the fluid-solid coupling mechanism in the high-efficiency case, attempting to analyze the reasons for its efficiency.

\begin{figure*}[htbp!]
    \centering
    \includegraphics[width=\textwidth]{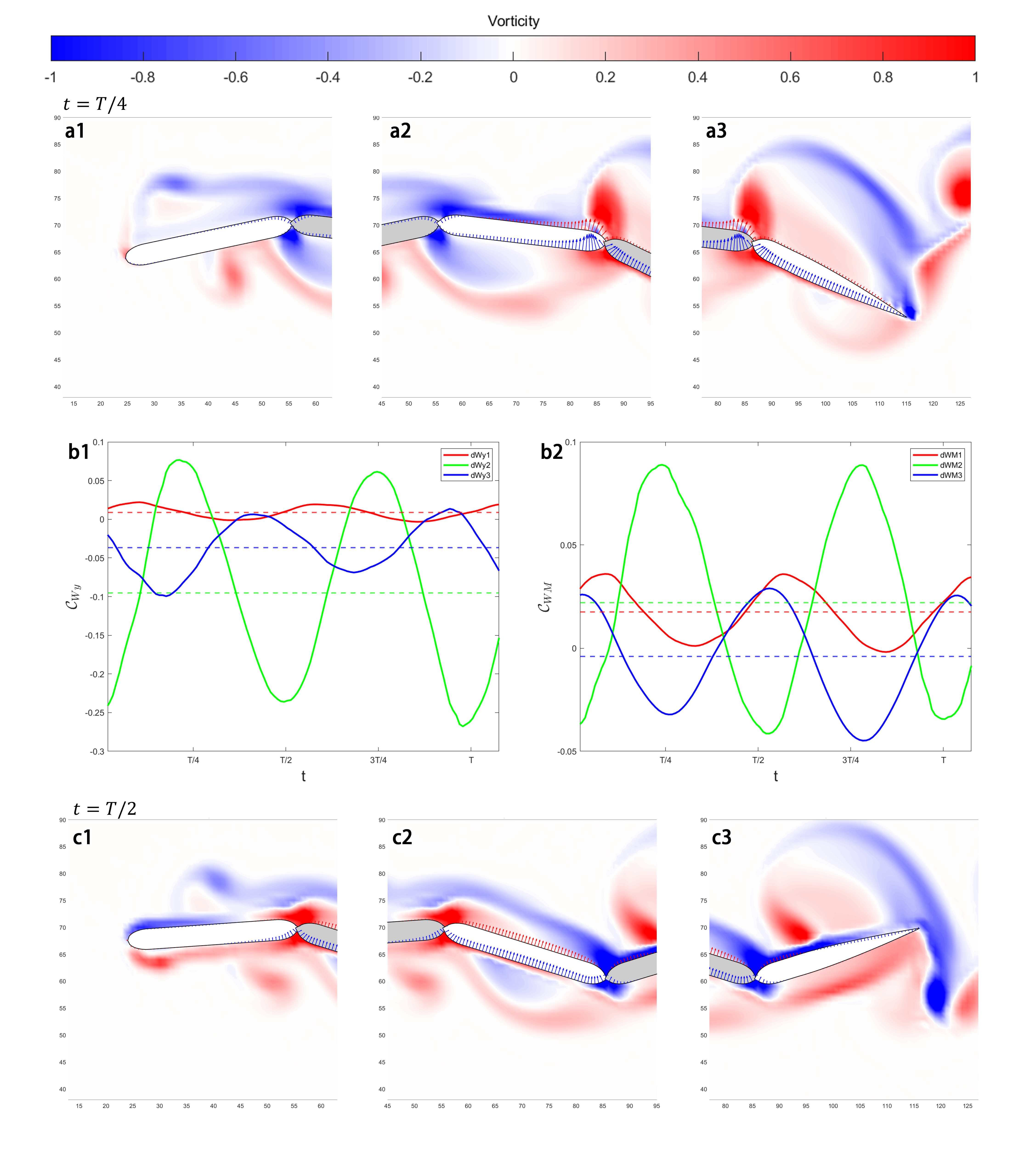}
    \caption{Vortex and pressure analysis of the flow field in the high-efficiency case. (a1-3)Vortex and pressure distribution of all three links at $t=T/4$ in a steady motion period. Note that the vortex data are shown in the flow field, while the magnitude and direction of the pressure distribution along the link surface are described in terms of arrow length and direction, where red represents positive pressure and blue represents negative. (b1-2)The energy absorbed by the structure from the flow in one motion period. Different colors represent the three links of the fish body, with the two sub-figures illustrating the y-direction(translating) and z-direction(rotating) energy absorption. (c1-3)Vortex and pressure distribution of all three links at $t=T/2$ in a steady motion period. }
    \label{fig:ana}
\end{figure*}

We plotted the energy consumption of three links within a stable period, dividing the consumption at each link into x, y, and z \textcolor{black}{directions}, representing the rotation. They have a summative relationship with the total energy consumption. Notably, due to our numerical settings, the displacement in the x direction can almost be ignored, making the energy input in the y and z directions significant components of the total energy, which refer to $C_{W_y}$ and $C_{W_M}$, as shown in sub-figures \ref{fig:ana}(b1) and (b2), illustrating the $dW_y$ and $dW_M$ within a motion period. Notice that when the $dW$ value is less than 0, energy is transferred from the link to the flow field, while when $dW$ is greater than 0, the flow field inputs energy to the link. The average energy input within the overall cycle is marked with dashed lines in the figure. The movement in the y direction consumes significant energy (especially the second and third links). In contrast, energy is transferred from the fluid to the fish body during link rotation, especially in the case of the second link.

At the $T/4$ phase, both the second and third links' energy input peaks. The second link absorbs energy from the fluid in the y and z directions, while the third link is the opposite. We plotted the flow field at this moment, where the vortex information is displayed in the computational domain, and the arrows alongside the links represent the pressure distribution on the fish body surface, with red being positive outward and blue being negative inward. At this moment, the second link is swinging downward with the first connection point as the axis. Simultaneously, due to the vortex below the second link, a strong negative pressure area is generated; a positive pressure distribution appears on the upper surface of the second link. Overall, the fluid's pressure difference and movement direction are consistent, thus doing positive work to the link. We can also see that the pressure distribution increases towards the tail end; thus, the overall torque direction is clockwise (or inward perpendicular to the paper, consistent with the overall rotation around the centroid), also doing positive work. For the third link, the situation is different; at this moment, its front end is following the previous link's decline, and a stronger vortex above forms a negative pressure area, making the upper part negative pressure, with greater intensity than the lower part. The tail of the third link is rising at this moment, receiving fluid resistance, naturally forming upper positive pressure and lower negative pressure. It is noted that the flow does negative work under such pressure distribution to the third link in the y and z directions, meaning the energy flows from the structure to the flow field.

Similarly, an examination is conducted on another peak, specifically focusing on the situation pertaining to the $T/2$ phase. At this point, the fluid does negative work in both directions of the second link, while the third receives positive energy. The overall movement can be seen as the second link rotating upward while simultaneously driving the third link to swing upward. The strong pressure difference between the upper and lower surfaces of the second link is contrary to the movement direction, thus doing negative work. At the same time, due to the vortex generated by link rotation, the head pressure is smaller, making the overall torque clockwise, contrary to the rotation direction, causing the rotation motion to consume energy. The middle upper surface of the third link has an evident low-pressure area, as the front half before the vortex has positive pressure, and the part after it has negative pressure, consistent with the movement direction. Considering the distribution, the y-direction energy input is almost zero, showing a slightly positive effect. At the same time, the rotation-direction torque is counterclockwise, consistent with the movement direction, so the fluid does positive work in rotation. Compared with the thrust-emphasizing case, both links utilized the characteristics of vortex generation from the interaction with the preceding structure at different times, combining with suitable motion patterns to absorb the energy from the flow field to the structure, significantly improving efficiency.

It is worth noting that the evolution of such leading-edge vortices is related to previous research on the hydrodynamics of flexible or deformable bodies. In the work of Hua et al.\cite{hua2013locomotion} on the flexible plates hydrodynamics, specific combinations of stiffness and frequency were found to enhance the overall propulsion efficiency compared to rigid bodies. Due to the elasticity of the flexible plate, the generation and rearward propagation of shedding vortices from the leading edge, combines with the trailing edge at the appropriate phase lag, resulting in energy saving and improved efficiency. In the analysis of biological fish and the design of biomimetic robotic fish by Liu et al.\cite{liu2010biological}, the amplitude and phase of joint motions were also parameterized to mimic the motion modalities of fish, yielding better locomotion performance. Compared to previous works, our reinforcement learning process not only contributes to understanding the underlying mechanisms for efficient swimming strategies, but also generates fine-grained motion strategies beyond periodicity motion, maximizing positive fluid-structure interactions to enhance overall swimming performance.

\section{Conclusion}\label{sec:conclusion}

This study leverages deep reinforcement learning (DRL) to discover the control scheme of a multi-joint bio-inspired robot. To enhance the performance of DRL, our approach encapsulates several pivotal methodologies: incorporating flow pressure sensing into the learning process that significantly extends the perception of hydrodynamic; adopting the transformer network architectures to suit the intricacies of the task; introducing the transferring of trigonometrical swimming policies' knowledge to expedite DRL training.

Interestingly, the optimization results of DRL are not singular but point toward different advantageous tendencies, including motion patterns with high thrust, high efficiency, and balanced characteristics. In the high efficiency scenario which we focus on, the fish exhibits high efficiency while maintaining a slightly positive thrust in the incoming flow, which can be considered an elegant cruising state. Analysis of each link reveals that the motion patterns learned through DRL allow for an intelligent combination of vortices generated by the leading-edge motion with the movements in the mid and posterior parts of the fish body. As the vortices and low-pressure areas are transferred rearward, they coincide with the corresponding link swinging towards the approaching vortex. This significantly enhances the fish body's overall utilization of the flow field energy and leads to a high-efficiency result.

The experiment results and analysis show that our framework has great potential in enhancing flow utilization and motion optimization for multi-link bionic fish. These advancements surpass the Pareto frontier of traditional parametric movements, provide excellent and precise swimming strategies suitable for varying requirements, and offer theoretical guidance for subsequent potential bionic fish movement design.


\begin{acknowledgments}
This work was supported by the National Key Research and Development Program (2022YFC2805200), start-up funding from Westlake University under grant number 041030150118 and Scientific Research Funding Project of Westlake University under Grant No. 2021WUFP017.

\end{acknowledgments}

\section{Author Declarations}
\subsection{Conflict of Interests}
The authors have no conflicts to disclose.

\subsection{Ethical Approval}
No experiments on animal or human subjects were used for the preparation of the submitted manuscript.

\subsection{Author Contributions}
Xinyu Cui and Boai Sun contributed equally to this work and should be considered co-first authors.

\textbf{Xinyu Cui}: Conceptualization (equal); 
Formal Analysis; 
Methodology (equal); Software (equal); Visualization (equal); Validation (equal); Writing - original draft preparation.
\textbf{Boai Sun}: Conceptualization (equal); 
Formal Analysis; 
Methodology (equal); Software (equal); Visualization (equal); Validation (equal); Writing - original draft preparation. 
\textbf{Yi Zhu}: Supervision (equal); Writing - review \& editing (equal). 
\textbf{Ning Yang}: Supervision (equal); Writing - review \& editing (equal). 
\textbf{Haifeng Zhang}: Resources (equal); Supervision (equal); Writing - review \& editing (equal). 
\textbf{Jun Wang}: Supervision (equal); Writing - review \& editing (equal). 
\textbf{Weicheng Cui}: Funding Acquisition; Supervision (equal); Writing - review \& editing (equal).
\textbf{Dixia Fan}: Conceptualization (equal); Resources (equal); Methodology (equal); Supervision (equal); Writing - review \& editing (equal).

\section*{Data Availability Statement}

The data supporting this study's findings are available from the corresponding author upon reasonable request.

\appendix

\section{Validation of the BDIM algorithm}\label{app:BDIM}
\begin{figure}[t!]
    \centering
    \includegraphics[width=0.45\textwidth]{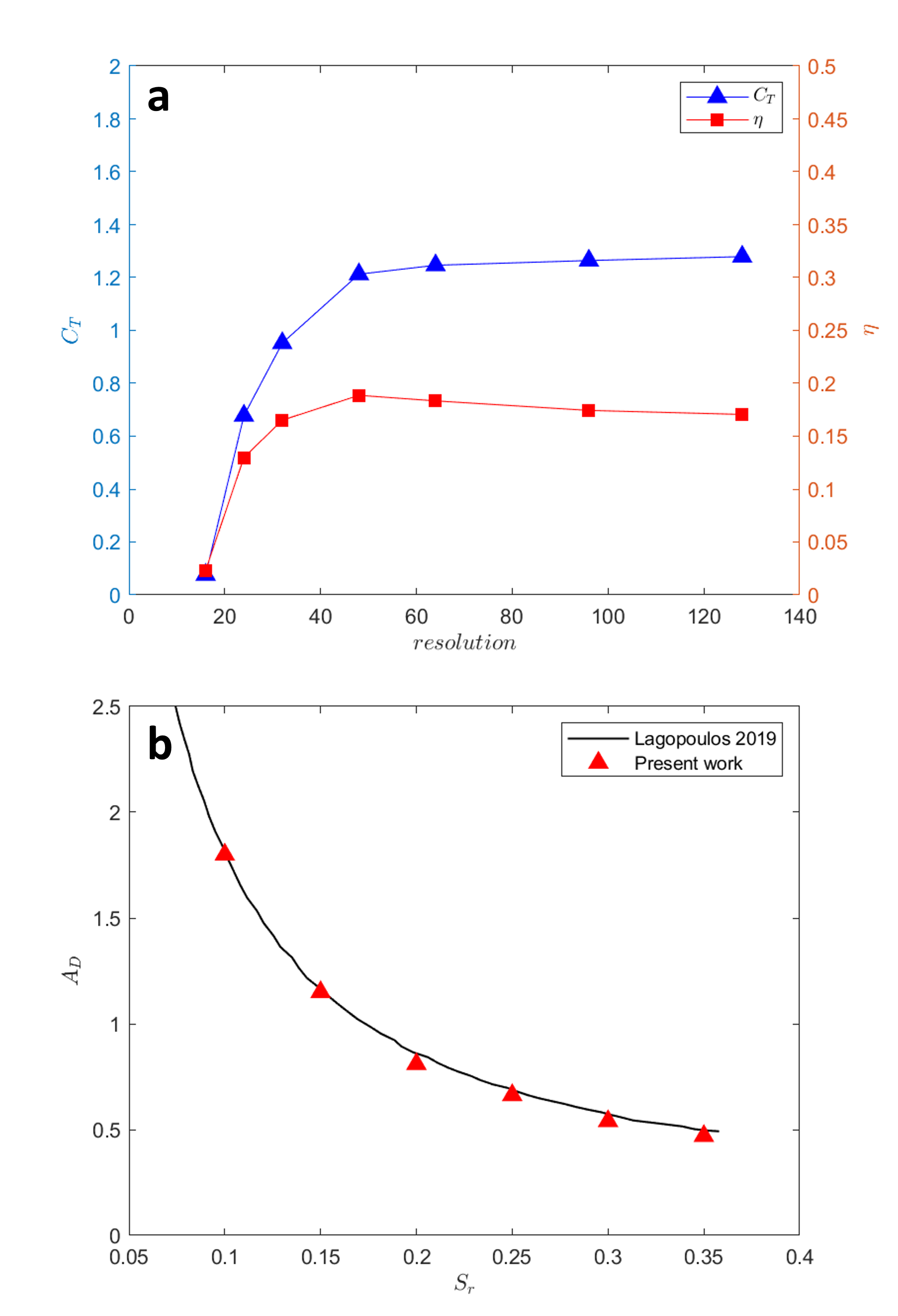}
    \caption{(a)$C_T$ and $\eta$ results under different resolutions of BDIM. (b)The validation of the BDIM algorithm, compared with Lagopoulos et al. 2019\cite{lagopoulos2019universal}. Note that the parameter settings are slightly different in Lagopoulos's work, and we cope with the previous study for convenience. The corresponding motion parameters are $Sr = St*C/L$, where $C$ is the chord thickness and $L$ is the chord length, and $A_D = 2y_C/C$, where $y_C$ represents the maximum y-directional displacement of the tail.}\label{fig:validation}
\end{figure}

To assess the convergence and accuracy of our algorithm, we select the work by Lagopoulos et.al in 2019\cite{lagopoulos2019universal} as the benchmark. In their study, a series of standard NACA0016 airfoil pitching experiments were conducted. We select the case of pitching motion around the leading edge at various frequencies, and attempt to replicate the amplitude when the thrust in the incoming flow is precisely zero. After getting various frequency-amplitude data points, we compare our data points with the curve in La's article. The convergence experiments shown in sub-figure \ref{fig:validation}(a) show that our experiments on thrust and efficiency from a grid density of 16 to 128 at the frequency and amplitude conditions of $Sr=0.3, A_D=2.0$ exhibit good convergence. In actual computations, our DRL environment requires the solver to provide results swiftly to generate substantial iterative data. Under such circumstances, we choose a grid density of 96 for extensive experiments to balance accuracy and efficiency, which we believe is reasonable. Compared with Lagopoulos's work in sub-figure \ref{fig:validation}(b), it is evident that the thrust-drag transition point provided by our algorithm aligns well with the curve, which confirms the validation of our BDIM algorithm.

\section{Penalty Functions}\label{app:penalty}

For simplicity, the penalty $\xi_n$ at step $n$ is expressed as $\xi$. 
Penalty $\xi$ is composed of 4 parts: $\xi = \xi_p + \xi_a + \xi_y + \xi_e$ ; penalty for unstable positions $\xi_p$, penalty for unstable angular acceleration $\xi_a$, penalty for unstable y-dimensional position $\xi_y$ ,and penalty for causing an early stopping: $\xi_e$.

\subsection{Penalty for Unstable Postures $\xi_p$}
\begin{equation}
\xi_p = 75 - max(|\theta_1|, 20) - max(|\theta_2|, 20) - max(|\theta_3|, 35) .
\end{equation}

In this equation, $\theta_1$ and $\theta_2$ represent the head's and trunk's angles relative to joint 0, while $\theta_3$ represents the tail's angle relative to joint 1. Penalty for unstable postures $\xi_p$ remains 0 when the link angles are in relative reasonable regions, which is $[-20, 20]$ for $\theta_1$ and $\theta_2$, and $[-35, 35]$ for $\theta_3$. 

\subsection{Penalty for Unstable Angular Acceleration $\xi_a$}
\begin{equation}
\xi_a = 10 * (0.15 - \sum_{i=1,2,3} max(|\ddot{\theta_i}|, 0.05)) .
\end{equation}

Similar to the penalty for unstable positions, $\ddot{\theta_i}$ represents each link's angular accelerations. Whenever the agent makes an attempt on actions making severe angular accelerations, it gets a penalty at each simulation step.

\subsection{Penalty for Unstable Y-Dimension Position $\xi_y$}
\begin{equation}
\xi_y = 5 - max(|y|, 5) .
\end{equation}

Whenever the agent moves out of the safe range in the y-direction set to $[-5, 5]$, it gets a penalty at each simulation step. This penalty is set to prevent the agent from continuously moving to one side.

\subsection{Penalty for Early Stopping $\xi_e$}
\begin{equation}
\xi_e =
\begin{cases}
  -100 & \text{Early Stopped;}  \\
  0 & \text{Otherwise. } 
\end{cases}
\end{equation}

Whenever the agent reaches the limit that the simulation needs to be early stopped in case of involving bad data to the training data, the simulation is terminated before the set episode length. Besides, the agent receives a relatively high penalty to avoid it from trying to reach these states.






\nocite{*}
\bibliography{aipsamp}

\end{document}